\documentclass{article}

   \usepackage[preprint,nonatbib]{neurips_2024}

\usepackage[utf8]{inputenc} 
\usepackage[T1]{fontenc}    
\usepackage{hyperref}       
\usepackage{url}            
\usepackage{booktabs}       
\usepackage{amsfonts}       
\usepackage{nicefrac}       
\usepackage{microtype}      
\usepackage{xcolor}         
\usepackage{enumitem}
\usepackage{amsmath} 

\usepackage{wrapfig}

\usepackage[textsize=tiny]{todonotes}
\usepackage{makecell}

\def\eg{\emph{e.g.}} 

\def\ie{\emph{i.e.}}

\definecolor{darkviolet}{rgb}{0.58, 0.2, 0.63}

\usepackage{xspace}
\newcommand{\model}{Director3D}
\newcommand{\modelname}{Director3D\xspace}

\title{\modelname: Real-world Camera Trajectory and 3D Scene Generation from Text}

\author{%
  Xinyang Li$^1$~
  Zhangyu Lai$^1$~
  Linning Xu$^3$~
  Yansong Qu$^1$
  \\\textbf{
  Liujuan Cao$^1$\thanks{Corresponding Author}~~
  Shengchuan Zhang$^1$~
  Bo Dai$^2$~
  Rongrong Ji$^1$
  }\\
  $^1$ Key Laboratory of Multimedia Trusted Perception and Efficient Computing,\\Ministry of Education of China, Xiamen University\\
  $^2$ Shanghai Artificial Intelligence Laboratory, $^3$ The Chinese University of Hong Kong
  \\
  \texttt{Code}: \url{https://github.com/imlixinyang/director3d}
}

\begin{document}

\maketitle

\begin{figure*}[!h]
  \centering
  \includegraphics[width=1\linewidth]{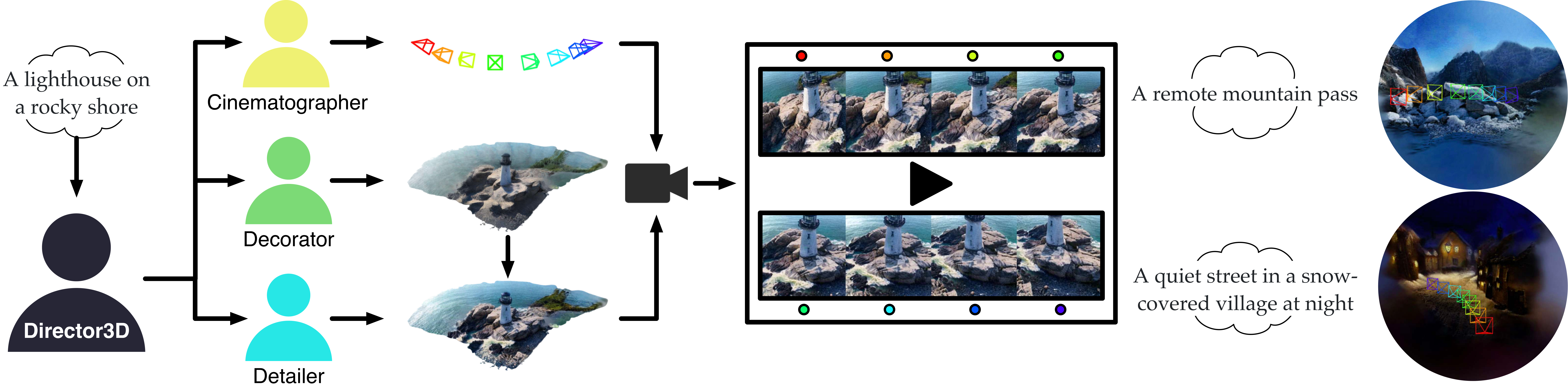}
  \caption{
  Given textual descriptions, \modelname employs three key components: the Cinematographer generates the camera trajectories, the Decorator creates the initial 3D scenes, and the Detailer refines the details. 
  }
  \label{fig.teaser}
\end{figure*}

\begin{abstract}
\label{abstract}
    Recent advancements in 3D generation have leveraged synthetic datasets with ground truth 3D assets and predefined cameras. However, the potential of adopting real-world datasets, which can produce significantly more realistic 3D scenes, remains largely unexplored.
    In this work, we delve into the key challenge of the complex and scene-specific camera trajectories found in real-world captures. 
    We introduce \textbf{\modelname}, a robust open-world text-to-3D generation framework, designed to generate both real-world 3D scenes and adaptive camera trajectories.
    To achieve this, (1) we first utilize a Trajectory Diffusion Transformer, acting as the \emph{Cinematographer}, to model the distribution of camera trajectories based on textual descriptions. 
    (2) Next, a Gaussian-driven Multi-view Latent Diffusion Model serves as the \emph{Decorator}, modeling the image sequence distribution given the camera trajectories and texts. This model, fine-tuned from a 2D diffusion model, directly generates pixel-aligned 3D Gaussians as an immediate 3D scene representation for consistent denoising.
    (3) Lastly, the 3D Gaussians are refined by a novel SDS++ loss as the \emph{Detailer}, which incorporates the prior of the 2D diffusion model.
    Extensive experiments demonstrate that \modelname outperforms existing methods, offering superior performance in real-world 3D generation.

\end{abstract}

\section{Introduction}
\label{intro}


Generating 3D scenes from texts holds great promise for industries such as gaming, robotics, and VR/AR. Previous methods~\cite{poole2022dreamfusion,sjc,shi2023mvdream}, which use score distillation sampling (SDS) to optimize 3D representations such as Neural Radiance Fields (NeRFs)~\cite{mildenhall2021nerf,muller2022instant}, involve lengthy and unstable optimization processes. 
In contrast, newer approaches employ feed-forward networks~\cite{hong2023lrm,xu2023dmv3d}, \eg, diffusion and reconstruction models, to directly generate 3D representations from text or text-guided multi-view images, significantly enhancing generation speed. Moreover, advancements in 3D Gaussian Splatting~\cite{kerbl3Dgaussians} further accelerate training and rendering speeds, driving the next wave of progress in text-to-3D generation~\cite{tang2023dreamgaussian,chen2024textto3dgsgen,yi2023gaussiandreamer,tang2024lgm,gslrm2024}.
However, most existing methods focus solely on object-level 3D generation. Recently, preliminary works~\cite{zhang2023text2nerf,hoellein2023text2room,chung2023luciddreamer,li2024dreamscene,zhou2024dreamscene360} have begun addressing scene-level 3D generation. Despite these efforts, visual quality, generation speed, and generalization remain suboptimal due to reliance on only 2D priors or limited few-classes 3D datasets.

In this work, we leverage real-world datasets (\eg, MVImgNet~\cite{yu2023mvimgnet} and DL3DV-10K~\cite{ling2023dl3dv}) to achieve realistic text-to-3D generation. 
However, real-world captures from in-the-wild scenes differ significantly from traditional object-level synthetic datasets, introducing new requirements for the text-to-3D generation framework.
Firstly, real-world captures feature complex, unpredictable, and scene-specific camera trajectories, unlike the controlled and predefined settings in object-level synthetic datasets like Objaverse~\cite{deitke2023objaverse} shown in Fig.~\ref{fig.1} (Left). 
Secondly, real-world scenes include unbounded backgrounds, complicating the use of common bounded 3D representations such as Tri-planes~\cite{chan2022efficient}. 
Lastly, the diversity and quantity of real-world captures are limited, potentially decreasing the generalization ability for open-world texts.

\begin{figure*}[!t]
  \centering
  \includegraphics[width=1\linewidth]{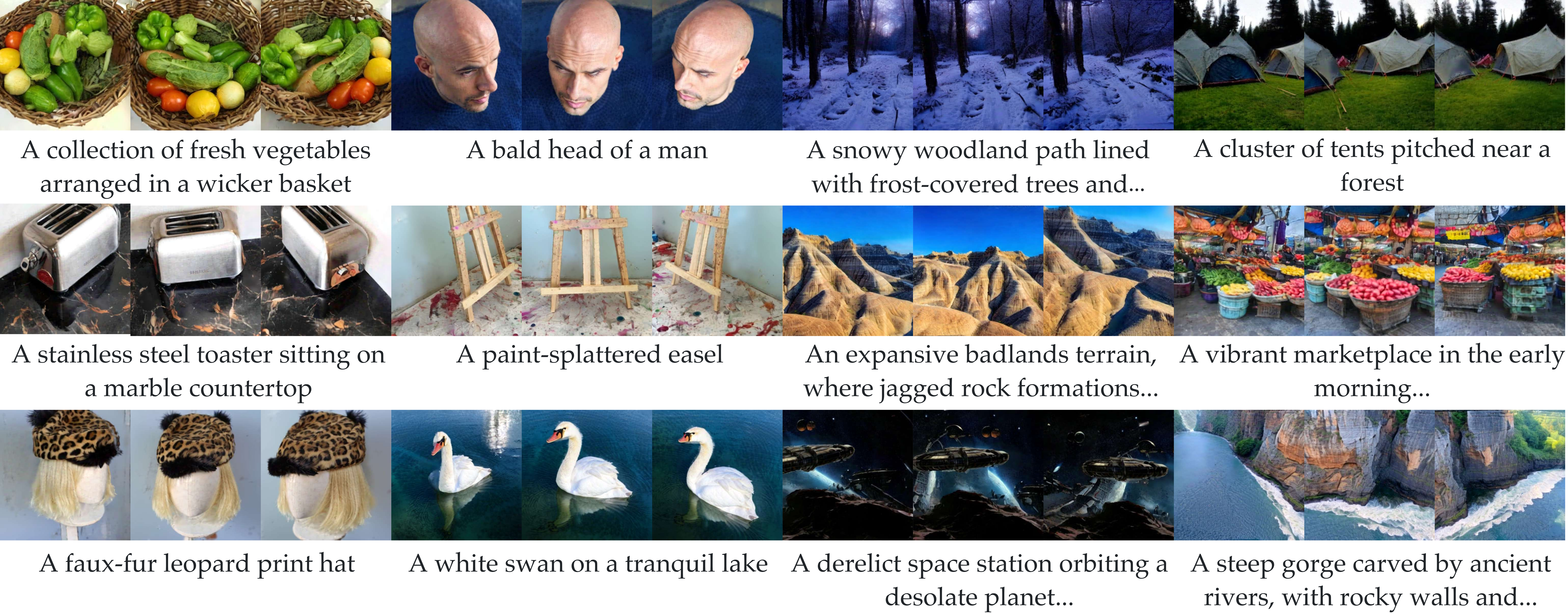}
  \vspace{-15pt}
  \caption{
  Multi-view image results rendered with the generated camera trajectories and 3D scenes. 
  %
  %
  }
  \label{fig.0}
\end{figure*}

We address these challenges with a novel framework, \modelname, illustrated in Fig.~\ref{fig.teaser} and Fig.~\ref{fig.1} (Right). 
Fig.~\ref{fig.0} shows that our framework supports generating 3D scenes across various domains.
In summary, our approach includes the following three key components:

$\bullet$  Traj-DiT (Trajectory Diffusion Transformer) as Cinematographer: Generates dense-view camera trajectories from text. Camera parameters (intrinsics and extrinsics) are treated as temporal tokens, and a Transformer model performs conditional denoising of the camera trajectory.

$\bullet$  GM-LDM (Gaussian-driven Multi-view Latent Diffusion Model) as Decorator: Uses a sparse-view subset of the camera trajectory for image sequence diffusion, generating pixel-aligned and unbounded 3D Gaussians as intermediate 3D representations. This model, fine-tuned from a 2D latent diffusion model, leverages strong priors and collaborative training with multi-view and single-view data to mitigate the limited diversity and quantity of real-world captures, enhancing generalization.

$\bullet$ SDS++ Loss as Detailer: Enhances visual quality of the 3D Gaussians by back-propagates a novel SDS++ loss from images rendered at randomly interpolated cameras within the trajectory.

\section{Related Works}


\begin{figure*}[!t]
  \centering
  \includegraphics[width=1\linewidth]{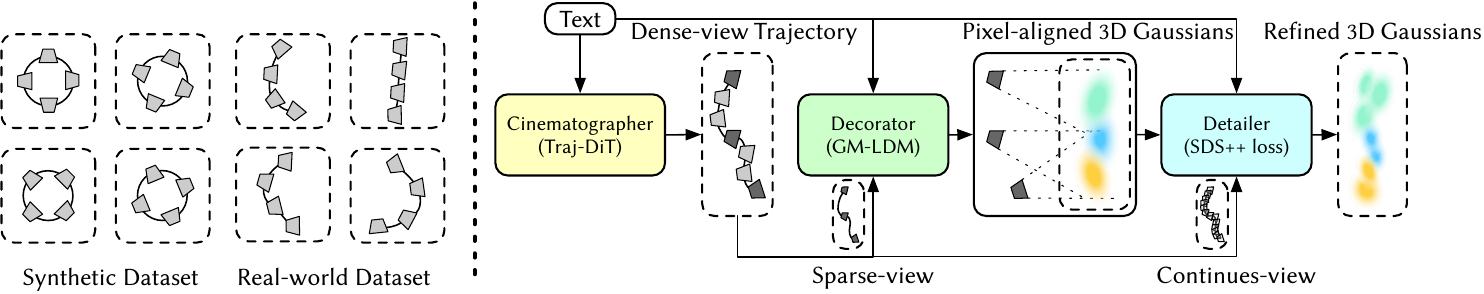}
  \caption{
  \textbf{Left}: Comparison of the simplified camera trajectory distributions between synthetic and real-world multi-view datasets. 
  \textbf{Right}: Pipeline and models of \model. 
  }
  \label{fig.1}
\end{figure*}

\noindent \textbf{Text-to-3D Generation.}
%
Since the introduction of DreamField~\cite{jain2022zero}, which combines the vision-language model CLIP~\cite{radford2021learning} with NeRFs~\cite{mildenhall2021nerf}, there has been substantial progress in the text-to-3D generation area.
DreamFusion~\cite{poole2022dreamfusion} and SJC~\cite{wang2023score} employ 2D image diffusion models to refine the 3D representation through Score Distillation Sampling (SDS).
Subsequent methods~\cite{wang2023prolificdreamer,katzir2023noise,zou2023sparse3d,bahmani20234d,wu2024hd} enhance the formulations to more effectively utilize 2D diffusion models for text-to-3D generation, achieving greater stability and visual quality.
Some methods~\cite{shi2023mvdream,liu2023unidream,long2023wonder3d,liu2023syncdreamer} suggest incorporating an additional 3D prior into 2D diffusion models to enhance 3D consistency. Additionally, various methods~\cite{nichol2022pointe,jun2023shape,tang2023volumediffusion,shue20233d,xu2023dmv3d,xu2024grm} employ 3D diffusion models to directly produce 3D assets.
However, most of these works primarily focus on object-level generation.
For scene-level 3D generation, preliminary works~\cite{hoellein2023text2room, zhang2023text2nerf, Lei_2023_rgbd2, SceneScape, Cohen-Bar_2023_setthescene, chung2023luciddreamer, shriram2024realmdreamer} combine image inpainting models~\cite{ho2020denoising,rombach2022high} and monocular depth estimation models~\cite{depthanything, midas} to progressively lift an image into a 3D scene with a user-defined camera trajectory.
Nonetheless, the multi-view inconsistency of image inpainting and monocular depth estimation can lead to undesirable geometry and texture artifacts.
Also, some works~\cite{tang2023mvdiffusion,ma2024fastscene} propose using panorama image diffusion models for 3D scene generation, which is limited to some specific types of scenes such as indoor rooms.
Therefore, developing a text-to-3D scene generation method with open-world generalization capabilities remains an unsolved problem.

\noindent \textbf{3D Gaussian Splatting (3DGS)}~\cite{kerbl3Dgaussians}
%
introduces parameterized 3D Gaussians as 3D representation and splatting-based rasterization technique for novel view synthesis based on dense-view images, significantly reducing the rendering time compared to NeRF-based methods~\cite{mildenhall2021nerf, wang2021neus, muller2022instant,barron2022mipnerf360,yu_and_fridovichkeil2021plenoxels}.
Current methods concentrate on improving the geometry quality~\cite{guedon2023sugar,yu2024gsdf, Huang2DGS2024}, stabilizing the training~\cite{scaffoldgs}, and adapting it to dynamic scene modeling~\cite{wu20234dgaussians, li2023spacetime,yang2023gs4d,jiang2023hifi4g}.
Further, to train a generalizable sparse-view reconstruction model with 3D Gaussians as the intermediate representation, some methods~\cite{chen2024mvsplat, charatan23pixelsplat, szymanowicz24splatter, xu2024grm, tang2024lgm, gslrm2024} propose to convert image features into pixel-aligned 3D Gaussians and optimize reconstruction models through back-propagating losses from rendered images.

Meanwhile, DreamGaussian employs 3DGS for text-to-3D generation, with SDS loss as the optimization objective.
GaussianDreamer~\cite{yi2023gaussiandreamer} and GSGen~\cite{chen2024textto3dgsgen} further enhance the generation quality and 3D consistency by initializing the 3D Gaussians based on point cloud diffusion models, Point-E~\cite{nichol2022pointe}.
The generalizable sparse-view reconstruction model (\eg, GRM~\cite{xu2024grm}, LGM~\cite{tang2024lgm}, and GS-LRM~\cite{gslrm2024}) for 3D Gaussians can also facilitate text-to-3D generation, using multi-view diffusion models~\cite{shi2023mvdream, li2024instant3d} to acquire sparse-view images as inputs.
Specifically, GS-LRM takes a step forward by employing a large video generation model (i.e., SORA~\cite{sora}) for text-to-3D scene generation.
However, these works heavily rely on the 3D consistency of the multi-view images, which the 2D-based diffusion models can not guarantee.
%
Distinctively, our GM-LDM employs pixel-aligned 3D Gaussians as the intermediate 3D representation for rendering-based multi-view diffusion, directly enforcing 3D consistency during the diffusion process and producing 3D representations.


\section{Preliminary}

\noindent \textbf{Latent Diffusion Models (LDMs)}~\cite{rombach2022high,saharia2022photorealistic}
consist of two key components: an auto-encoder~\cite{kingma2022autoencoding} and a latent denoising network.
The autoencoder establishes a bi-directional mapping from the space of the original data to a low-resolution latent space:
$
z = \mathcal{E}(x), x = \mathcal{D}(z),
$
where $\mathcal{E}$ and $\mathcal{D}$ are the encoder and decoder, respectively. 
The latent denoising network $\epsilon_\theta$ is trained to denoise noisy latent given a specific timestep $t$ and condition $y$. 
Its training objective for $\epsilon$-prediction is defined as:
\begin{equation}
L=\mathbb{E}_{
x,\epsilon\sim\mathcal{N}(0,1),t}
\Big[\|\epsilon-\epsilon_\theta(z_t,y,t)\|_2^2\Big],
\label{eq.2}
\end{equation}
where the noisy latent is obtained by $z_t = \sqrt{\bar{\alpha}_t} E(x) + \sqrt{1 - \bar{\alpha}_t} \epsilon$, $\bar{\alpha}_t$ is a monotonically decreasing noise schedule and $\epsilon \sim \mathcal{N}(0,1)$ is a random noise.
During inference, a random noise is sampled as $z_T \sim \mathcal{N}(0,1)$. 
By continuously denoising the random noise $z_T$ with condition $y$ (\eg, text embedding), we can derive a fully denoised latent $\hat{z}$.
Then, the denoised latent $\hat{z}$ is fed into the latent decoder $\mathcal{D}$ to generate the high-resolution image $\hat{x} = \mathcal{D}(\hat{z})$.

\noindent \textbf{Multi-view Diffusion Models} aim to model the distribution of multi-view images $\mathcal{X}$ with 3D consistency, where each image is captured by a distinct camera within the same static 3D scene.
Its objective for $x_0$-prediction can be written as:
\begin{equation}
L=\mathbb{E}_{
\mathcal{X},\epsilon\sim\mathcal{N}(0,1),t}
\Big[\|\mathcal{X} - 
\mathcal{X}_\theta(\mathcal{X}_t,\mathcal{C},y,t))\|_2^2\Big],
\label{eq.3}
\end{equation}
where $\mathcal{C}$ represents the camera parameters for the different views.
Early works~\cite{shi2023mvdream, liu2023unidream} in this field are based on 2D LDMs. 
They fine-tune the 2D LDMs by integrating cross-view connections between the multi-view images into the original single-view 2D LDMs, using multi-view data rendered from 3D datasets. 
These methods lack strict 3D consistency since there is no actual 3D representation during multi-view denoising.
%
%
A more advanced approach, DMV3D~\cite{xu2023dmv3d}, employs a 3D reconstruction model to generate noise-free 3D representations and predict multi-view images from noisy multi-view inputs by a rendering-based denoising process.
This enables 3D generation tasks to be accomplished without per-asset optimization during inference.
%

\noindent \textbf{Score Distillation Sampling (SDS)}~\cite{poole2022dreamfusion,sjc}
uses a pretrained 2D diffusion model to optimize 3D representation.
Considering a differentiable 3D representation parameterized by $\mathcal{G}$
and a rendering function denoted as $\mathcal{R}$, the rendered image produced for a given
camera pose $c$ can be expressed as $x=\mathcal{R}(\mathcal{G}, c)$. SDS distills the prior of a 2D LDM to optimize 3D representation $\mathcal{G}$ as follows:
\begin{equation}
\nabla_{\mathcal{G}} \mathcal{L}_{\mathrm{SDS}}=\mathbb{E}_{t, \epsilon, c}\left[w(t)\left(\hat{\epsilon}-\epsilon\right) \frac{\partial \mathcal{E}(\mathcal{R}(\mathcal{G}, c))}{\partial \mathcal{G}}\right]
\label{eq.4}
\end{equation}
where $\epsilon$ is the ground truth noise, $\hat{\epsilon}$ is the noise predicted by the 2D LDM with $z_t$ as input for timestep $t$, and $w(t)$ represents a weighting function that varies according to the timestep $t$.
The SDS loss can be also converted into a reconstruction-like objective~\cite{zhu2023hifa}:
\begin{equation}
\mathcal{L}_{\mathrm{SDS}}=\mathbb{E}_{t, \epsilon, c}\left[w(t) \frac{\sqrt{\bar{\alpha}_t}}{\sqrt{1 - \bar{\alpha}_t}}\|z - \hat{z}\|_2^2 \right], \text{where } \hat{z} = (z_t - \sqrt{1 - \bar{\alpha}_t} \hat{\epsilon}) / \sqrt{\bar{\alpha}_t}
\label{eq.5}
\end{equation}

\section{Method}
\label{sec.4}

\subsection{Problem Formulation and Overview of \modelname}
We consider the multi-view dataset of real-world captures as a joint distribution of image sequences and camera trajectories conditioned on texts, denoted as $p((\mathcal{X}, \mathcal{C})|y)$. Here, $\mathcal{X}=\{x_i\}_{i=1}^M$ represents the image sequence, $\mathcal{C}=\{c_i\}_{i=1}^M$ denotes the camera trajectory, and $M$ is the number of views. 
To model this joint distribution, we separately handle the conditional distributions $p(\mathcal{C}|y)$ and $p(\mathcal{X}|(\mathcal{C},y))$ (see Appendix~\ref{app.discussions} for detailed discussions). Furthermore, we model each image in the sequence as a rendered view of a unified 3D scene representation $\mathcal{G}$ under the corresponding camera, expressed as $x_i=\mathcal{R}(\mathcal{G}, c_i)$, where $\mathcal{R}$ is the 3D rendering function.

\modelname addresses this by incorporating three collaborative processes analogous to roles in film production: the Cinematographer, the Decorator, and the Detailer.
Firstly, the Trajectory Diffusion Transformer (Traj-DiT), serving as the Cinematographer, models the distribution of dense-view camera trajectories, as detailed in Sec.~\ref{sec.4.2}.
For image sequences, directly modeling the dense-view distribution is complex and resource-intensive. To address this, we use a Gaussians-driven Multi-view Latent Diffusion Model (GM-LDM), acting as the Decorator, to model the image distribution through a sparse subset of dense views. This model utilizes pixel-aligned 3D Gaussians as the intermediate representation, described in Sec.~\ref{sec.4.3}.
Finally, to improve the visual quality of the generated 3D scenes, we employ a novel SDS++ loss, functioning as the Detailer, to refine the 3D Gaussians through dense-camera interpolation rendering, as presented in Sec.~\ref{sec.4.4}.

\begin{figure*}[!t]
  \centering
  \includegraphics[width=1\linewidth]{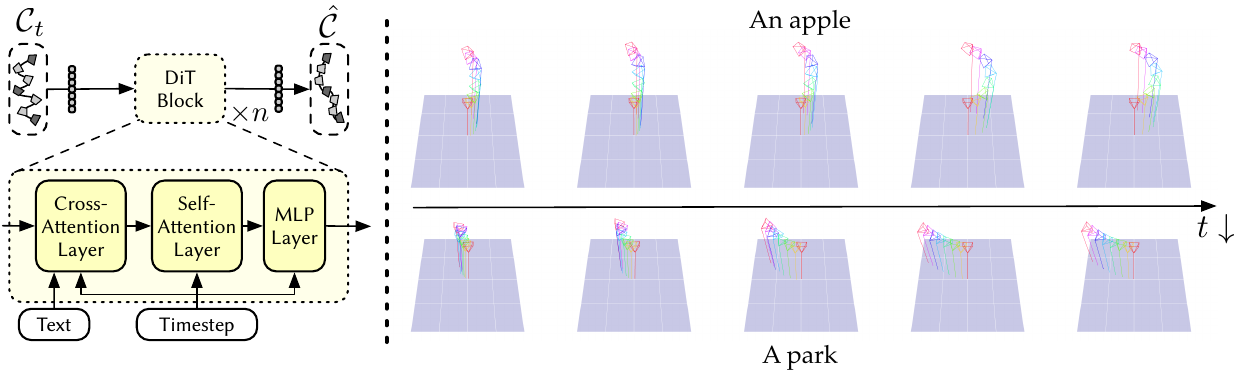}
  \caption{
  \textbf{Left}: Architecture of Traj-DiT. 
  \textbf{Right}: Visualization of the predicted camera trajectory for different denoising timesteps.
  }
  \label{fig.2}
\end{figure*}

\subsection{Traj-DiT as Cinematographer}
\label{sec.4.2}

%
To model the trajectory distribution, we represent the camera trajectory $\mathcal{C}$ as a set of camera parameters $c_i = \{\mathbf{r}_i, \mathbf{t}_i, \mathbf{f}_i, \mathbf{p}_i\}$, where $\mathbf{r} \sim \mathrm{SO}(3)$ and $\mathbf{t} \sim \mathbb{R}^3$ are the rotation and translation of the camera poses, $\mathbf{f} \sim \mathbb{R}_+^2 $ is the focal lengths and $\mathbf{p} \sim \mathbb{R}^2$ is the principle points.
To ensure consistency and comparability across scenes, we normalize the trajectory for each scene in two steps: First, we convert all camera poses to be relative to the first one so that the first camera pose is an identity matrix; Then, we re-scale the translation to make the distance from the first to the farthest camera to $1$.
We adapt the architecture of the Diffusion Transformer (DiT)~\cite{peebles2023dit} to generate camera trajectories, as illustrated in Fig.~\ref{fig.2} (Left). The temporal order of real-world captures, akin to video sequences, necessitates a learnable temporal embedding to differentiate between cameras of different frames. This embedding helps the model capture the sequential dependencies inherent in real-world data.
Each DiT block includes a cross-attention layer to extract information from text embeddings encoded by the CLIP~\cite{radford2021learning} text encoder. Additionally, the timestep $t$ modulates the pre-layer normalization and post-layer output scalar, similar to the original DiT, allowing the model to learn temporal dynamics effectively.
%
The model is trained to minimize the $x_0$-prediction diffusion objective:
\begin{equation}
L=\mathbb{E}_{
\mathcal{C},\epsilon\sim\mathcal{N}(0,1),t}
\Big[\|\mathcal{C} - 
\mathcal{C}_\theta(\mathcal{C}_t,y,t)\|_2^2\Big],
\label{eq.6}
\end{equation}
where $\mathcal{C}_\theta$ is the parameterized Traj-DiT model and $\mathcal{C}_t$ is the noisy camera trajectory.
%

By leveraging the strengths of the DiT architecture, we aim to enhance the fidelity and coherence of the generated trajectories, instead of relying on pre-defined ones. We showcase two examples of the predicted camera trajectories for different denoising steps $t$ in Fig.~\ref{fig.2} (Right), demonstrating the effectiveness of our model in generating smooth and accurate camera paths.
%

\subsection{GM-LDM as Decorator}
\label{sec.4.3}


We propose GM-LDM to model the image sequence distribution \( p(\mathcal{X}|(\mathcal{C}, y)) \) and generate immediate 3D Gaussians as the joint 3D scene representation. The GM-LDM, fine-tuned from the Stable Diffusion model with a slightly modified architecture, leverages its image generation prior to enhance 3D scene generation.
For efficiency, diffusion is applied to a sparse-view subset of the camera trajectory, significantly reducing computational overhead. 
During training, sparse-view images are processed through the frozen latent encoder \(\mathcal{E}\) to obtain multi-view latents \(\mathcal{Z} \sim \mathbb{R}^{N \times c \times h \times w}\), where \(N \leq M\). 
Noise is then added to these multi-view latents \(\mathcal{Z}\) to produce noisy latents \(\mathcal{Z}_t\).
%
%

%
\paragraph{2D-based Denoising.}
The noisy multi-view latents \(\mathcal{Z}_t\) are fed into the latent denoising network \(Z_\theta\) in parallel. For convenience, we modify the \(\epsilon\)-prediction of the original Stable Diffusion model to \(x_0\)-prediction. The denoised multi-view latents are obtained as \(\{\hat{\mathcal{Z}}, \mathcal{F}\} = Z_\theta(\{\mathcal{Z}_t\}, \mathcal{C}', y, t)\), where \(\mathcal{F}\) represents additional multi-view features for enhanced 3D information.
Sparse-view cameras \(\mathcal{C}'\) are integrated into the network by combining the ray-maps \((\boldsymbol{o} \times \boldsymbol{d}, \boldsymbol{d})\)~\cite{xu2023dmv3d} with the noisy latents, where \(\boldsymbol{o}\) and \(\boldsymbol{d}\) denote the origin and direction of pixel-aligned rays, respectively. We replace self-attention blocks in the original 2D latent denoising network with cross-view self-attention blocks~\cite{shi2023mvdream} to better capture multi-view correlations.
The denoised multi-view latents \(\hat{\mathcal{Z}}\) are supervised using a simple multi-view latent diffusion objective:
\begin{equation}
\mathcal{L}_{\text{2d}}=\mathbb{E}_{
\mathcal{X},c,y,\epsilon,t}
\Big[\|\mathcal{Z} - \hat{\mathcal{Z}}\|^2_2\Big].
\label{eq.7}
\end{equation}

\begin{figure*}[!t]
  \centering
  \includegraphics[width=1\linewidth]{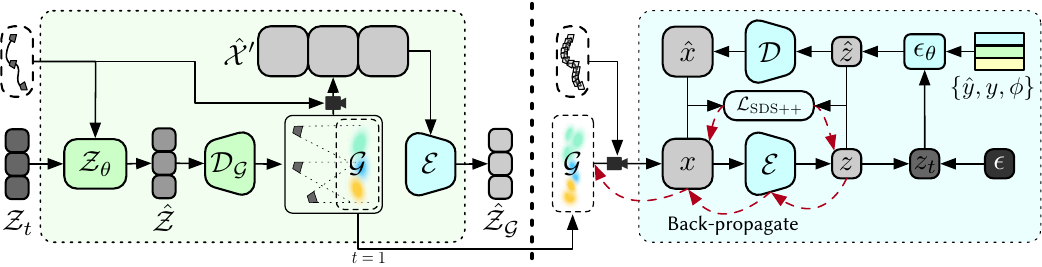}
  \caption{
  \textbf{Left}: Architecture of GM-LDM. The model is fine-tuned from a 2D LDM with minor modifications, performing rendering-based denoising for generating initial 3D Gaussians.
  \textbf{Right}: Pipeline of calculating SDS++ loss, which refines the 3D Gaussians with the original 2D LDM.
  }
  \label{fig.3}
\end{figure*}

\textbf{Rendering-based Denoising.}
%
To generate 3D Gaussians for rendering-based denoising, the denoised multi-view latents $\hat{\mathcal{Z}}$ and additional features $\mathcal{F}$ are input into a Gaussians decoder $\mathcal{D}_\mathcal{G}$. This decoder outputs Gaussian features $\{\tau_i, \boldsymbol{q}_i, \boldsymbol{s}_i, \alpha_i, \boldsymbol{c}_i\} = \mathcal{D}_\mathcal{G}(\hat{\mathcal{Z}}_i, \mathcal{F}_i)$, where $\tau_i$, $\boldsymbol{q}_i$, $\boldsymbol{s}_i$, $\alpha_i$, and $\boldsymbol{c}_i$ represent the depth, rotation quaternion, scale, opacity, and spherical harmonics coefficients of $256\times 256$ 3D Gaussians for view $i$, respectively.
The Gaussians decoder $\mathcal{D}_\mathcal{G}$ is initialized with the weights of the original latent decoder $D$, with re-initialized first and last convolutional layers to handle the additional features and specific Gaussian channels. The predicted depth is then converted into pixel-aligned Gaussian positions $\boldsymbol{\mu}_i = \boldsymbol{o}_i + \tau_i \boldsymbol{d}_i$.
The multi-view 3D Gaussians $\mathcal{G} = \{\boldsymbol{\mu}_i, \boldsymbol{q}_i, \boldsymbol{s}_i, \alpha_i, \boldsymbol{c}_i\}_{i=1}^N$ are concatenated to jointly represent the 3D scene. During training, views are randomly sampled from the dense-view camera trajectory to supervise the predicted 3D Gaussians in image space, ensuring consistent and accurate 3D scene representation:
\begin{equation}
\mathcal{L}_{\text{3d}}=\mathbb{E}_{x,c,y,\epsilon,t}
\Big[ \ell(x, 
\mathcal{R}(\mathcal{G}, c)) \Big],
\label{eq.8}
\end{equation}
where $\mathcal{R}$ is the rendering function, $\ell(\cdot,\cdot)$ is a reconstruction loss penalizing the difference between images, and $(x, c) \in (\mathcal{X}, \mathcal{C})$ is the ground truth of an image and camera pair from the dense views.
We use a combination of MSE loss and LPIPS~\cite{zhang2018unreasonable} loss for the reconstruction loss $\ell$, similar to the original reconstruction loss of the original auto-encoder.
%
%
The total training loss is simply the sum of the above losses:
$\mathcal{L}=\mathcal{L}_{\text{2d}} + \mathcal{L}_{\text{3d}}$.
This approach leverages the strengths of multi-view data and pixel-aligned Gaussian representations to enhance the fidelity and coherence of the generated 3D scenes.
By fine-tuning from a robust 2D LDM and using a sparse-view subset, we strike a balance between performance, efficiency, and generalizability, enabling the generation of coherent 3D scenes. 
%
As shown in Fig.~\ref{fig.3} (Left), during inference, images are rendered with the input cameras $\mathcal{C}'$ and encoded by the latent encoder $\mathcal{E}$ to obtain Gaussian-driven denoised latents:
\begin{equation}
\hat{\mathcal{Z}}_\mathcal{G} = \mathcal{E}(\hat{\mathcal{X}}'), \text{where } \hat{\mathcal{X}}' = \mathcal{R}(\mathcal{G}, \mathcal{C}').
\end{equation}
Inspired by Dual3D~\cite{li2024dual3d}, GM-LDM inference can toggle between 2D-based and rendering-based denoising. The 2D-based denoising offers better generalization, aligning closely with the original Stable Diffusion, while rendering-based denoising ensures superior 3D consistency due to its immediate joint 3D representation. The 3D Gaussians generated in the final denoising step serve as the initial 3D scene for subsequent refinement.
%

%
Collecting and annotating real-world multi-view datasets is laborious, often resulting in limited diversity and quantity, which hinders generalization for open-world texts. To address this, we follow MVDream~\cite{shi2023mvdream} and collaboratively train the GM-LDM using both multi-view and 2D datasets to enhance generalization.
We treat single-view images as a special case of multi-view images with $N=M=1$ and apply the same rendering process and training losses, which increases the diversity of training data, thereby improving the model's ability to generalize across diverse scenarios.

\subsection{SDS++ Loss as Detailer}

\begin{figure*}[!t]
  \centering
  \includegraphics[width=1\linewidth]{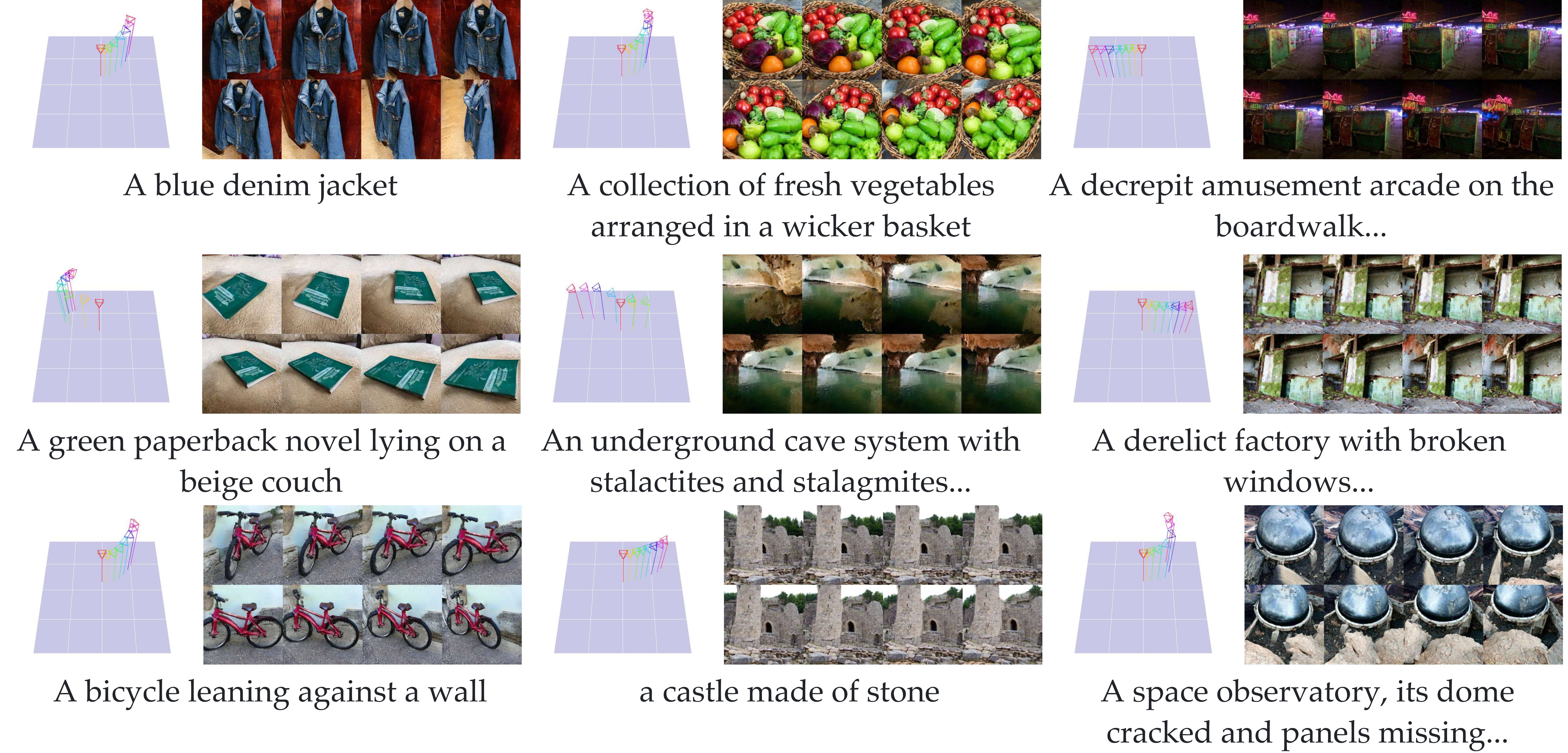}
  \caption{
  Generation results of \modelname for both camera trajectories and image sequences.
  }
  \label{fig.6}
\end{figure*}

\label{sec.4.4}

%
To enhance the details and visual quality of the 3D Gaussians, we propose the SDS++ loss, leveraging the 2D diffusion prior for refinement. Our research on existing SDS-based methods identifies three key points crucial for success, this includes:
(1) \textbf{Appropriate Target Distribution}~\cite{wang2023prolificdreamer}: This ensures that the rendered images align effectively with the textual conditions, avoiding over-smoothing and over-saturation.
(2) \textbf{Adaptive Estimation of the Current Distribution}~\cite{wang2023prolificdreamer,yang2023lods}: This provides a counter optimization objective, pushing the rendered images away from the current distribution to enhance details.
(3) \textbf{Latent-space and Image-space Objectives}~\cite{zhu2023hifa}: Combining these objectives helps prevent noisy or over-smoothing artifacts that may arise from using only one of them.
During refining, we first render the 3D Gaussians $\mathcal{G}$ with a randomly sampled camera $c$ from the continuous interpolated camera trajectory to produce an image $x = \mathcal{R}(\mathcal{G}, c)$. This image is encoded into the latent space by $\mathcal{E}(x) = z$, then disturbed with randomly sampled noise and timestep $t$ to produce a noisy latent $z_t$. The 2D diffusion model $\epsilon_{\theta}$ then predicts the denoised latent $\hat{z}$ from $z_t$.
As illustrated in Fig.~\ref{fig.3} (Right), the proposed SDS++ loss can be formulated by:
\begin{equation} 
\mathcal{L}_{\text{SDS++}} = \mathbb{E}_{t,c,\epsilon}
\left[w(t) \frac{\sqrt{\bar{\alpha}_t}}{\sqrt{1 - \bar{\alpha}_t}}
\left(\lambda_z \| z-\hat{z}\|^2_2+\lambda_x\|x-\hat{x}\|^2_2\right)\right],
\end{equation}
where $\lambda_z$ and $\lambda_x$ are the weights for latent-space and image-space objectives, respectively, $\hat{z}$ is the predicted latent, and $x = \mathcal{D}(\hat{z})$ is the predicted image.
The predicted latent can be derived by Eq.~\ref{eq.5}.
We use a compositional predictions $\hat{\epsilon}$ as follows:
\begin{equation} 
\begin{aligned} 
\hat{\epsilon} = \hat{\epsilon}_{\text{trg}} - \hat{\epsilon}_{\text{src}} + \epsilon.
\end{aligned}
\end{equation}
%
Instead of setting $\hat{\epsilon}_{\text{src}} = \epsilon$ as in the standard SDS loss, we introduce a learnable source prediction $\hat{\epsilon}_{\text{src}}$ for adaptive estimation of the current distribution:
\begin{equation} 
\hat{\epsilon}_{\text{src}} = \epsilon_\theta(z_t, \hat{y}, t),
\end{equation}
where $\epsilon_\theta(z_t, \hat{y}, t)$ uses a learnable text embedding $\hat{y}$ to efficiently estimate the current distribution. This approach leverages the original latent denoising network~\cite{yang2023lods} and is trained by minimizing
$
\|\epsilon_\theta(z_t, \hat{y}, t) - \epsilon\|^2_2
$
along with the refining process.
%
The target prediction employs classifier-free guidance for improved text alignment:
\begin{equation} 
\begin{aligned} 
\hat{\epsilon}_{\text{trg}} = \omega_{\text{cfg}} \cdot (\epsilon_\theta(z_t,y,t) - \epsilon_\theta(z_t,\phi,t)) + \epsilon_\theta(z_t,\phi,t),
\end{aligned}
\end{equation}
%
where $\omega_{\text{cfg}}$ is the classifier-free guidance scale.
%
%
SDS++ loss integrates the above three key points, ensuring efficient and realistic refinement of 3D Gaussians.

%

\section{Experiments}

\subsection{Implementation Details}

\label{sec.5.1}


We utilize the MVImgNet~\cite{yu2023mvimgnet} for object-level and DL3DV10K~\cite{ling2023dl3dv} for scene-level real-world multi-view datasets. Text prompts for each scene are generated using the multi-modal large language model InternLM-XComposer~\cite{internlmxcomposer}. To enhance generalization, we incorporate the 2D dataset LAION~\cite{schuhmann2022laion}.
For GM-LDM, we set the lengths of dense and sparse views to $M=29$ and $N=8$, respectively. The classifier-free guidance scale is $7.5$ for 2D-based denoising and $1$ for rendering-based denoising to ensure 3D consistency. Following Dual3D~\cite{li2024dual3d}, we balance 3D consistency and generalization by using 1/10 rendering-based denoising steps.
Image and latent resolutions are set to $256$ and $32$, respectively. For SDS++ loss, the weights for latent-space and image-space losses are $\lambda_{z} = 1$ and $\lambda_{x} = 0.01$. $\omega_{\text{cfg}}$ is set to $7.5$, with refining iterations set to $1000$. Generating a scene takes approximately 5 minutes.
%
%
Further implementation details are provided in Appendix~\ref{app.implemental_details}.

\subsection{Generation Results.}


We show the generation results of camera trajectories and image sequences for various text prompts in Fig.~\ref{fig.6}. 
For object-level prompts, the generated camera trajectories typically circle and face the objects, aligning well with the distribution in MVImgNet. In contrast, scene-level prompts yield more diverse and complex camera trajectories, showcasing the effectiveness of our Traj-DiT model.
Our method also generates realistic images across different types of prompts, demonstrating the effectiveness and generalization ability of GM-LDM and SDS++ loss. Additional generation results are available in Appendix~\ref{app.more_generation_results}.
These results highlight the robustness of our approach in handling both object-level and scene-level prompts for 3D scene generation.

\begin{figure*}[!t]
  \centering
  \includegraphics[width=1\linewidth]{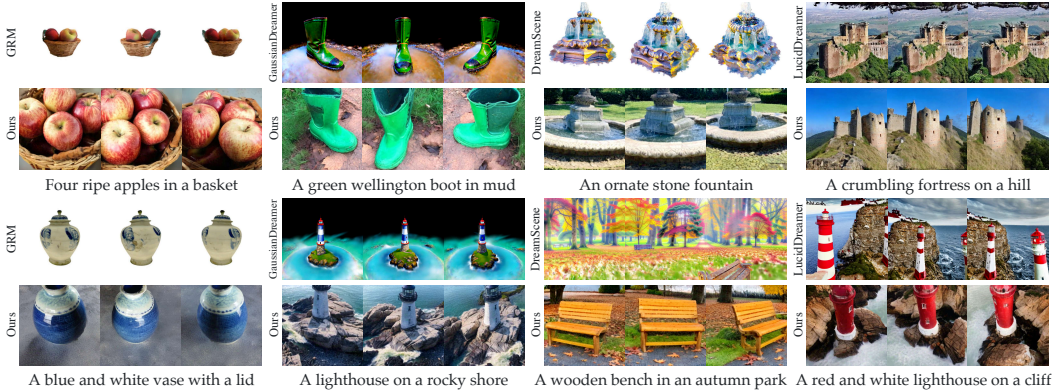}
  \caption{
  Qualitative comparison between \modelname and different baselines.
  }
  \label{fig.4}
\end{figure*}

\subsection{Qualitative Comparison.}

We qualitatively compare our method with several baseline methods, as shown in Fig.~\ref{fig.4}.
(1) GRM~\cite{xu2024grm} is a feed-forward text-to-3D generation method for 3D Gaussians using multi-view images generated by 2D-based diffusion models. It supports only object-level 3D generation. Our comparisons using object descriptions show that GRM produces unrealistic 3D Gaussians limited to objects, due to its training on synthetic datasets. In contrast, our method generates high-quality 3D scenes with both objects and backgrounds.
(2) GaussianDreamer~\cite{yi2023gaussiandreamer} is a state-of-the-art SDS-based method for 3D Gaussians, integrating priors from both 2D and 3D diffusion models. While it can generate objects with ground layers, it tends to produce over-saturated textures. Our method, however, generates more realistic scenes with better handling of shadows, lighting, and material reflections.
(3) DreamScene~\cite{li2024dreamscene} uses Formation Pattern Sampling and strategic camera sampling for 3D Gaussians. Since it is not open-sourced, we use examples from its project page. Although it generates scene-wide consistent 3D scenes, the results are overly saturated and cartoonish.
(4) LucidDreamer~\cite{chung2023luciddreamer}, based on Text2NeRF~\cite{zhang2023text2nerf}, uses 2D foundation models for 3D Gaussians. While it can generate photo-realistic textures, the multi-view consistency is poor, with visible artifacts at object edges due to inaccurate monocular depth estimation. It also struggles with excessive object generation from descriptive prompts due to reliance on single-view inpainting.
These comparisons highlight the superior performance of the proposed \modelname for realistic 3D generation.

\subsection{Quantitative Comparison.}
\begin{table}[!t]
    \centering
    \caption{Quantitative comparison of different models with text prompts in T3Bench.}
    \begin{tabular}{l|c|c|c}
    \toprule
      Method   & \makecell{BRISQUE~$\downarrow$}  & \makecell{NIQE~$\downarrow$}& \makecell{CLIP-Score~$\uparrow$} \\
      \midrule
        DreamFusion~\cite{poole2022dreamfusion} & 90.2 & 10.48  & 67.4 \\
        Magic3D~\cite{lin2023magic3d} & 92.8 & 11.20  & 72.3 \\
        LatentNerf~\cite{metzer2022latent} & 88.6 & 9.19  & 68.1 \\
        SJC~\cite{sjc} & 82.0 & 10.15  & 61.5 \\ 
        Fantasia3D~\cite{fantasia3d} & 69.6 & 7.65  & 66.6 \\ 
        ProlificDreamer~\cite{wang2023prolificdreamer} & 61.5 & 7.07  & 69.4 \\ 
        \midrule
        Ours \textit{w/o} refining & 37.1 & 6.41  & 80.0 \\ 
        \midrule
        Ours & \textbf{32.3} & \textbf{4.35} & \textbf{85.5} \\ 

        \bottomrule
    \end{tabular}
    \label{tab.1}
\end{table}

We present a quantitative comparison between our framework and several baseline models in Tab.~\ref{tab.1}. 
For this experiment, we use the Single-Object-with-Surroundings\footnote{\url{https://github.com/THU-LYJ-Lab/T3Bench/blob/main/data/prompt_surr.txt}} set of T3Bench~\cite{he2023t3bench}, which contains 100 prompts closely matching the descriptions in MVImgNet. The quantitative results are evaluated using CLIP-Score~\cite{hessel2021clipscore}, NIQE~\cite{niqe}, and BRISQUE~\cite{brisque} metrics.
For each 3D scene generated by different methods, we render a video and uniformly sample 36 frames to calculate the average score for each metric. For baselines without adaptive camera trajectories, videos are rendered by circling around the 3D representations at a fixed elevation.
The BRISQUE and NIQE results indicate that our method significantly outperforms existing baseline models in terms of image quality. Additionally, the CLIP-Score shows our method's superior ability to align generated images with their textual descriptions, even without refining.
These results underscore the robustness and effectiveness of our framework in generating high-quality, semantically aligned 3D scenes.
%


\subsection{Ablation Study}




\begin{wrapfigure}{r}{0pt}
\centering
\includegraphics[width=0.38\linewidth]{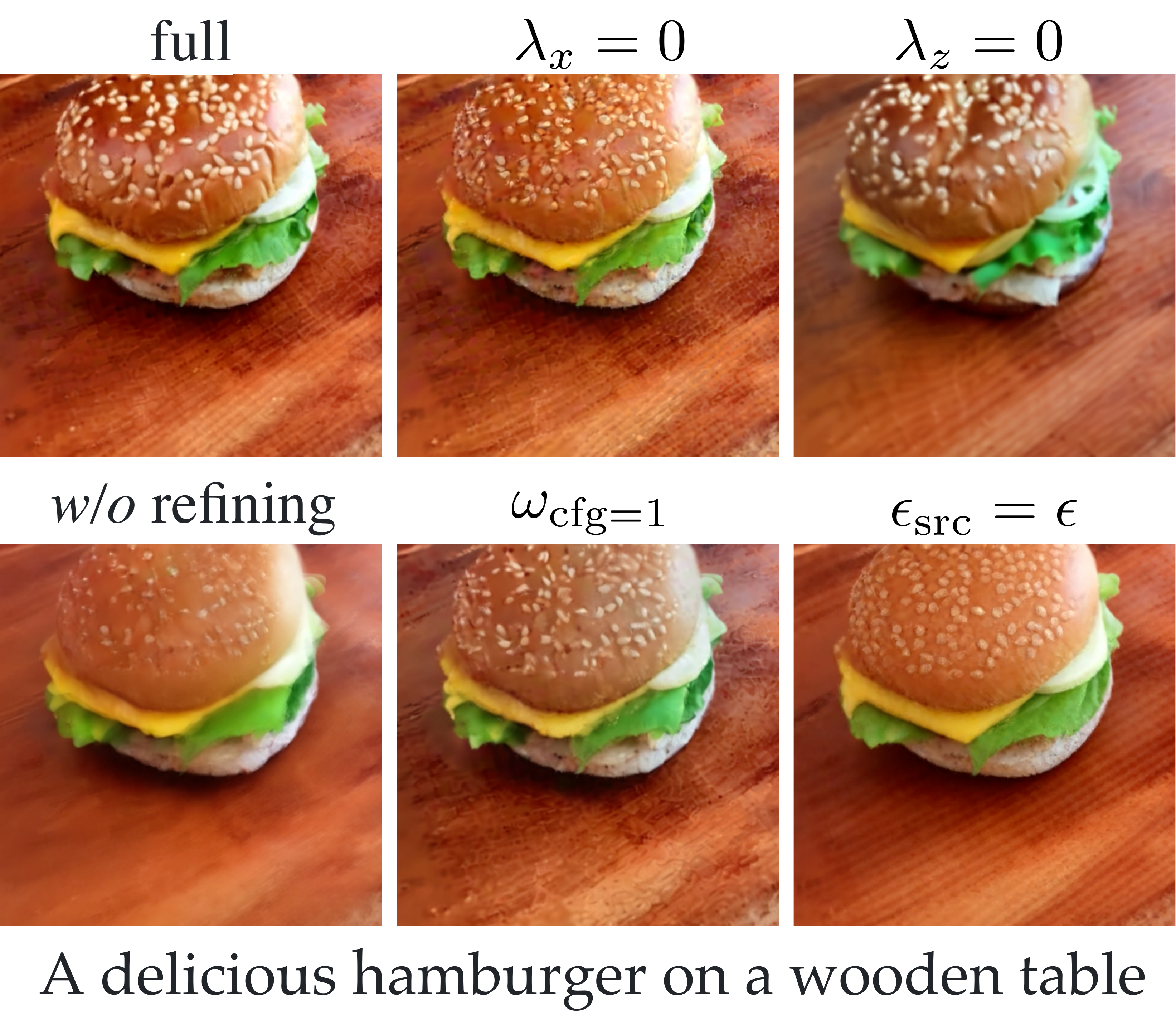}
\vspace{-50pt}
\caption{Ablation of SDS++ loss}
\label{fig.5}
\end{wrapfigure}

\modelname is formulated to a sequential process to model the joint distribution of camera trajectories and image sequences. The Traj-DiT component generates the base camera trajectory for each scene, while the GM-LDM provides the initial 3D scene. These two models are crucial for producing meaningful 3D scenes.
Our ablation study primarily focuses on the SDS++ loss to highlight its significance. 
As shown in Fig.~\ref{fig.5}, we conduct comprehensive experiments to analyze the impact of the SDS++ loss.
First, we remove the refining process entirely. Although the initial 3D Gaussians generated by GM-LDM match the input text, the visual quality is unsatisfactory, with missing details. This is expected due to the limited diversity and quantity of the multi-view dataset used for training GM-LDM.
Setting $\epsilon_{\text{src}} = \epsilon$ degrades the SDS++ loss into SDS+ loss~\cite{zhu2023hifa}. The results show a significant decrease in visual quality and an over-smoothing issue, highlighting the importance of adaptive estimation of the current distribution.
Setting $\omega_{\text{cfg}}=1$ turns the SDS++ loss into the LODS loss~\cite{yang2023lods} with an additional image-space objective. This results in noisy details, as the conditional noise prediction alone does not provide a clear optimization direction.
By setting $\lambda_{x} = 0$ and $\lambda_{z} = 0$ respectively, we observe that using only the latent-space objective leads to noisy details and artifacts, while using only the image-space objective results in missing details.
The full model, incorporating the proposed SDS++ loss, achieves the best visual quality with clear and realistic details. These findings underscore the importance of each component in the SDS++ loss and its role in refining 3D Gaussians.
Additional ablation studies, including those with randomly generated trajectories, are provided in Appendix~\ref{app.traj}.

\section{Conclusion}

In this paper, we propose an open-world text-to-3D generation framework capable of generating real-world 3D scenes with adaptive camera trajectory, named \modelname.
We first introduce a Cinematographer (\ie, Traj-DiT) that can generate dense-view camera trajectories given texts.
Then, a Decorator (\ie, GM-LDM) and a Detailer (\ie, SDS++ loss) are proposed for initial generation and further refining, respectively, with 3D Gaussians as the 3D scene representation.
%
We demonstrate the effectiveness of our method with extensive experiments.
We believe our work makes essential contributions to the text-to-3D generation community, especially in discovering the potential of leveraging real-world multi-view datasets for realistic 3D generation.
Our future works include improving the generation scope, boosting model efficiency and quality, and leveraging more datasets.

\section{Acknowledgements}
This work was supported by National Science and Technology Major Project (No. 2022ZD0118202), the National Science Fund for Distinguished Young Scholars (No.62025603), the National Natural Science Foundation of China (No. U21B2037, No. U22B2051, No. 62176222, No. 62176223, No. 62176226, No. 62072386, No. 62072387, No. 62072389, No. 62002305 and No. 62272401), and the Natural Science Foundation of Fujian Province of China (No.2021J01002,  No.2022J06001).


{\small
\bibliographystyle{unsrt}
\bibliography{arxiv.bib}
}

\clearpage
\newpage

\appendix


\section{Implemental Details}
\label{app.implemental_details}

\noindent \textbf{Datasets.}
For real-world multi-view datasets, we leverage MVImgNet~\cite{yu2023mvimgnet} and DL3DV10K~\cite{ling2023dl3dv} datasets for object-level and scene-level captures, respectively.
The MVImgNet dataset is a large-scale object-level dataset, comprising 219,188 videos across 238 classes\footnote{\url{https://github.com/GAP-LAB-CUHK-SZ/MVImgNet}, under the MVImgNet Terms of Use}. 
%
%
The DL3DV10K dataset consists of $10$K scene-level videos with various indoor, outdoor, urban, and suburban environments\footnote{\url{https://github.com/DL3DV-10K/Dataset}, under the DL3DV-10K Terms of Use}. 
%
%
%

\noindent \textbf{Traj-DiT setup.}
Our Traj-DiT is trained with the Adam optimizer~\cite{KingBa15}, a constant learning rate of $1e^{-4}$ and $(\beta_1, \beta_2) = (0.9, 0.95)$. 
The batch size is set to $256$. 
%
%
The number of DiT blocks is $8$ and the hidden size is $512$.
Training takes about $2$ days with $1$ NVIDIA Tesla A100 GPUs for $50$K iterations.
%
We use $1000$ steps during training and reduce it to $100$ steps with DDIM~\cite{song2022denoising} during inference.

\noindent \textbf{GM-LDM setup.}
Our GM-LDM is trained with the Adam optimizer, a constant learning rate of $5e^{-5}$ and $(\beta_1, \beta_2) = (0.9, 0.95)$. 
%
%
The batch size is also set to $128$. 
Training takes about a week with $16$ NVIDIA Tesla A100 GPUs for $150$K iterations.
We use Stable Diffusion v2.1\footnote{\url{https://huggingface.co/stabilityai/stable-diffusion-2-1-base}} as our initial model.
The noise schedule and other hyperparameters are consistent with the Traj-DiT.
%

\noindent \textbf{SDS++ loss setup.}
For refining, the learning rates of the Gaussian parameters $\boldsymbol{\mu}$, $\boldsymbol{q}$, $\boldsymbol{s}$, $\alpha$, and $\boldsymbol{c}$ are set to be $0.0001$, $0.01$, $0.001$, $ 0.01$, and $0.01$, respectively.
%
%
The learning rate of the learnable text embedding $\hat{y}$ is set to $0.001$.
%
%
The timestep $t$ is annealing from $0.75T$ to $0.02T$ with a square root schedule, where $T$ is the total denoising steps.
We use Stable Diffusion v2.1 as our refining model as well.
The rendering resolution is set to $512$ during refinement.

\section{Discussions}

\noindent
\textbf{Ways to Model $p(\mathcal{X}, \mathcal{C})$.}
\label{app.discussions}
There are different approaches to modeling the joint distribution of image sequences and camera trajectories for real-world captures.
We choose to independently model the distribution of camera trajectories and the conditional distribution of image sequences based on the camera trajectories.
The image sequence is directly rendered with a joint 3D representation by the proposed GM-LDM, ensuring 3D consistency.
Another alternative is to first directly model the image sequences and then infer the camera trajectories from the image sequences.
Concurrent work, GS-LRM~\cite{gslrm2024}, adopts this approach. Specifically, it uses pose-free image sequences generated by SORA, then recovers camera information through COLMAP~\cite{schoenberger2016sfm}, and finally performs sparse-view 3D reconstruction. 
However, this approach heavily relies on the 3D consistency of the pose-free image sequences and the accurate estimation of the camera parameters.
One can also choose to directly model the joint distribution of camera trajectories and image sequences. However, this requires further exploration of network architectures and camera-differentiable 3DGS operator.
We will explore the potential of this approach in future work.

\noindent
\textbf{Why \modelname Needs a Detailer?}
The core reason remains the limitations of multi-view datasets in terms of diversity and quantity, leading to biased and sparse distributions. 
Although MVImgNet has a large number of scenes, it consists of only 238 categories, with the majority being simple objects staged in indoor scenes.
While DL3DV-10K exhibits good diversity, it only consists of 10,000 scenes in total.
Therefore, even under the 2D data and model priors, it is still challenging for GM-LDM to meet the quality requirements for directly generating realistic 3D scenes with details.
This limitation can be alleviated by introducing more diverse and larger multi-view datasets.

\noindent
\textbf{Limitations.}
\label{limitations}
Although the proposed \modelname possesses the capability to generate real-world 3D scenes with adaptive camera trajectories, it still has some limitations. 
First, the supported views of our GM-LDM are currently limited, which restricts the range of perspectives in the generated 3D scenes.
Further, because the GM-LDM is currently trained with only two real-world datasets, the reliance on an additional refining process for open-world generalization limits the efficiency of our framework, which may be alleviated by introducing a wider variety of datasets~\cite{reizenstein21co3d, zhou2018stereo, dai2017scannet}.
Similar to text-to-image diffusion models, the success rate of our model decreases when generating with complicated and compositional prompts, objects with exact numbers, and articulated objects.

\noindent
\textbf{Broader Impacts.}
This paper presents a framework whose goal is to advance the field of text-to-3D generation for efficiently generating realistic 3D scene assets.
Since it is a 3D generative framework, it has the potential
to create harmful 3D content if the user provides harmful text prompts as inputs.
This harmful behavior can be avoided by integrating harmful content detection models, as used in 2D generative models.

\section{Diversity and Fine-grained Control}

\begin{figure*}[!t]
  \centering
  \includegraphics[width=1\linewidth]{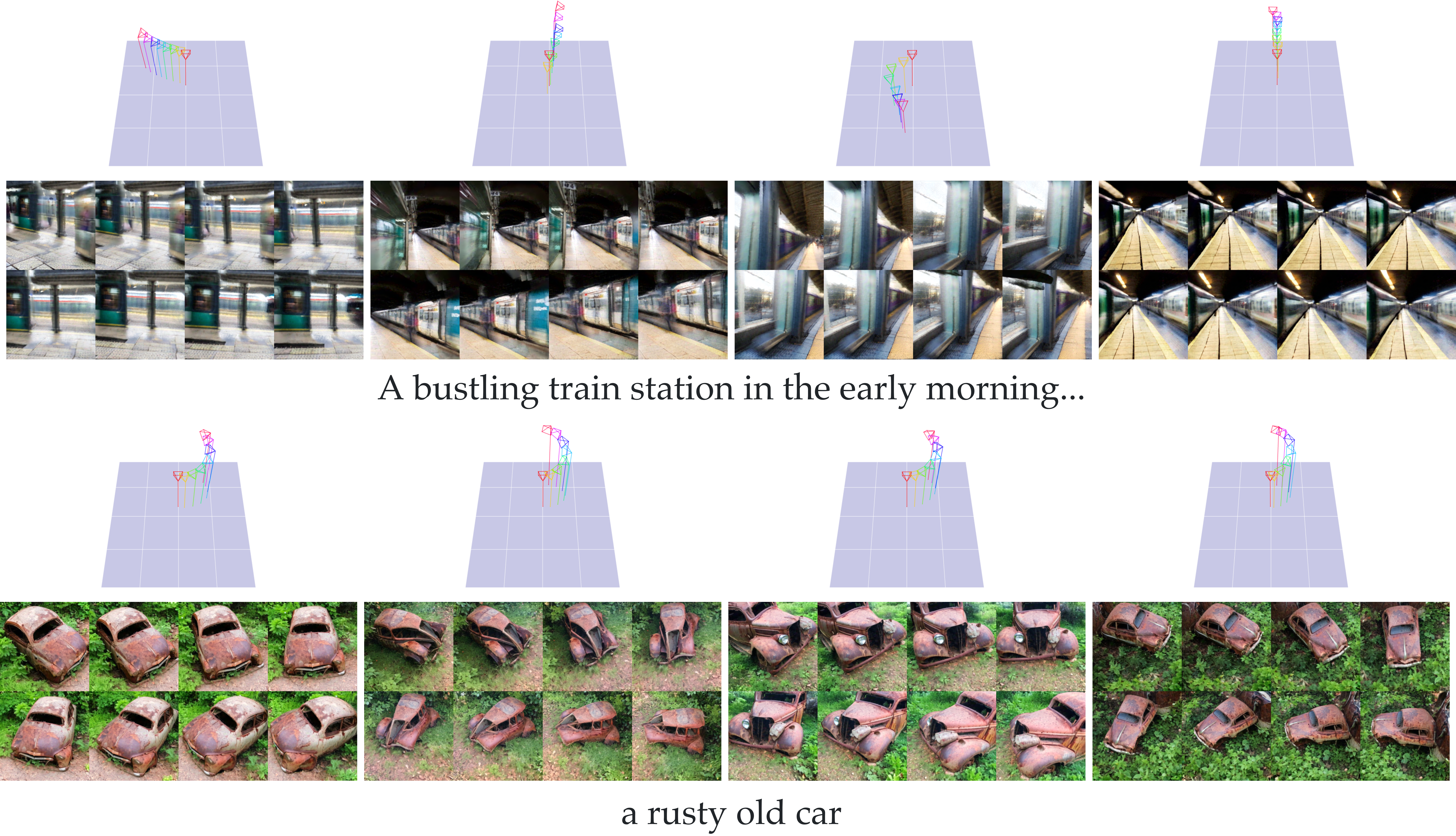}
  \caption{
  Generation results with diversity.
  }
  \label{fig.7}
\end{figure*}

We showcase the diverse generation results with the same prompts in Fig.~\ref{fig.7}.
The results show that \modelname is able to generate diverse camera trajectories and 3D scenes even with a single prompt.

\begin{figure*}[!t]
  \centering
  \includegraphics[width=1\linewidth]{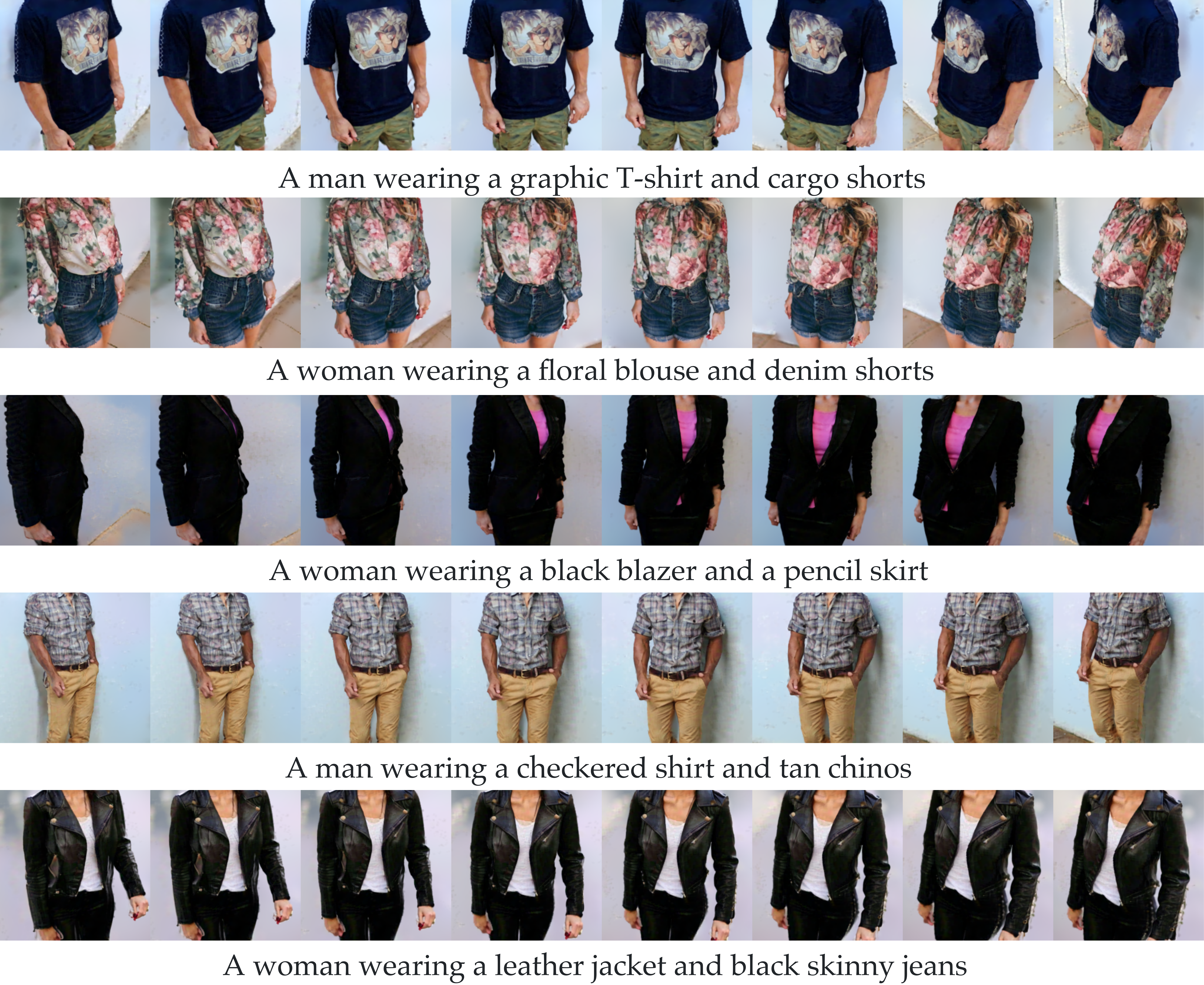}
  \caption{
  Generation results with fine-grained control.
  }
  \label{fig.8}
\end{figure*}

We showcase the body generation results with fine-grained control (\ie, gender, and clothing) in Fig.~\ref{fig.8}.
Note that there are no categories for human body in the multi-view datasets used for training the GM-LDM.
The results show that \modelname is able to finely control the 3D scenes with text modifications.

\section{More Generation Results}
\label{app.more_generation_results}

\begin{figure*}[!t]
  \centering
  \includegraphics[width=1\linewidth]{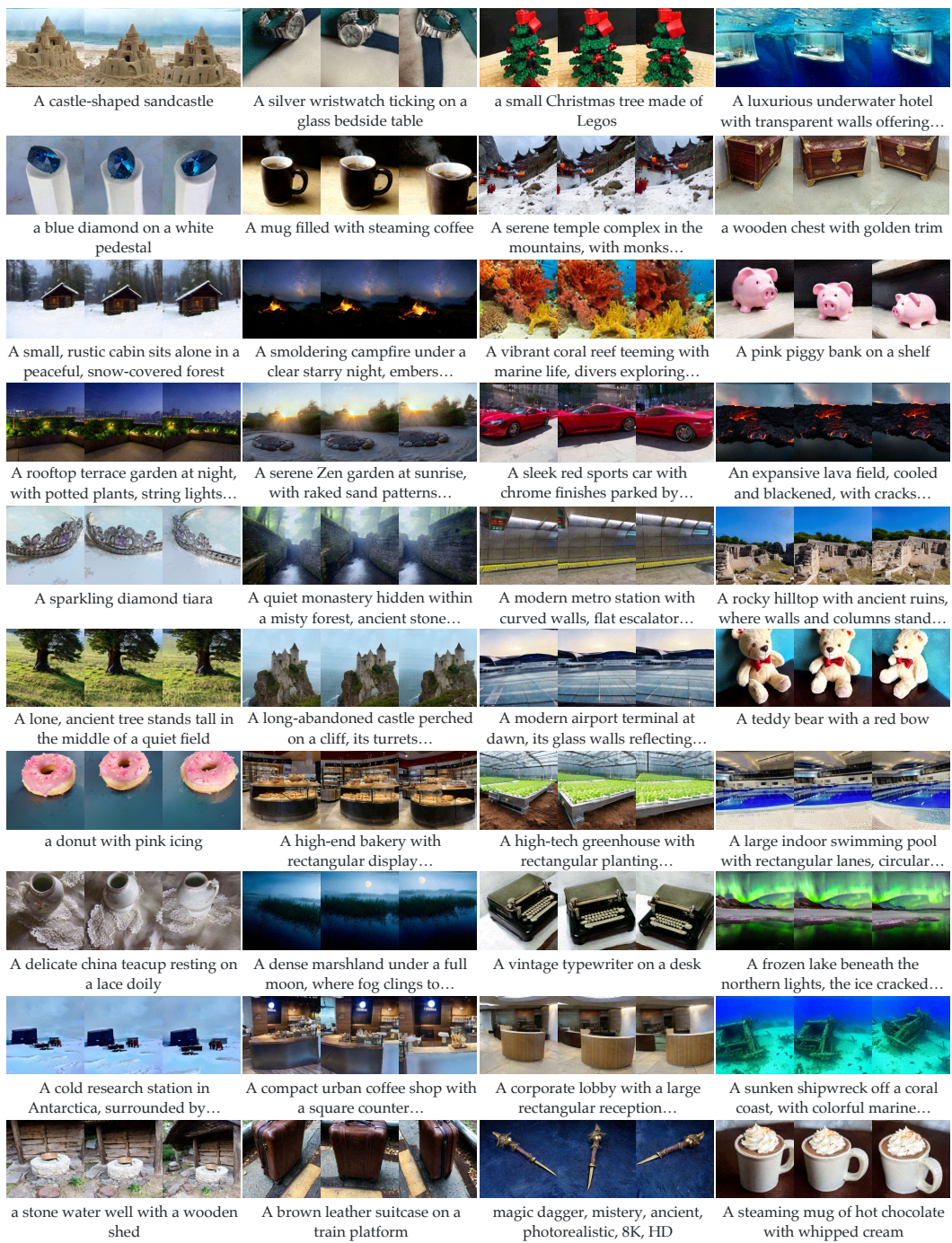}
  \caption{
  More multi-view image results.
  }
  \label{fig.9}
\end{figure*}

We further provide 40 cases of multi-view image results in Fig.~\ref{fig.9} generated from wide-range text prompts to demonstrate the visual quality and generalization ability of our method.

\section{Importance of Scene-specific Trajectories}
\label{app.traj}

\begin{figure*}[!t]
  \centering
  \includegraphics[width=1\linewidth]{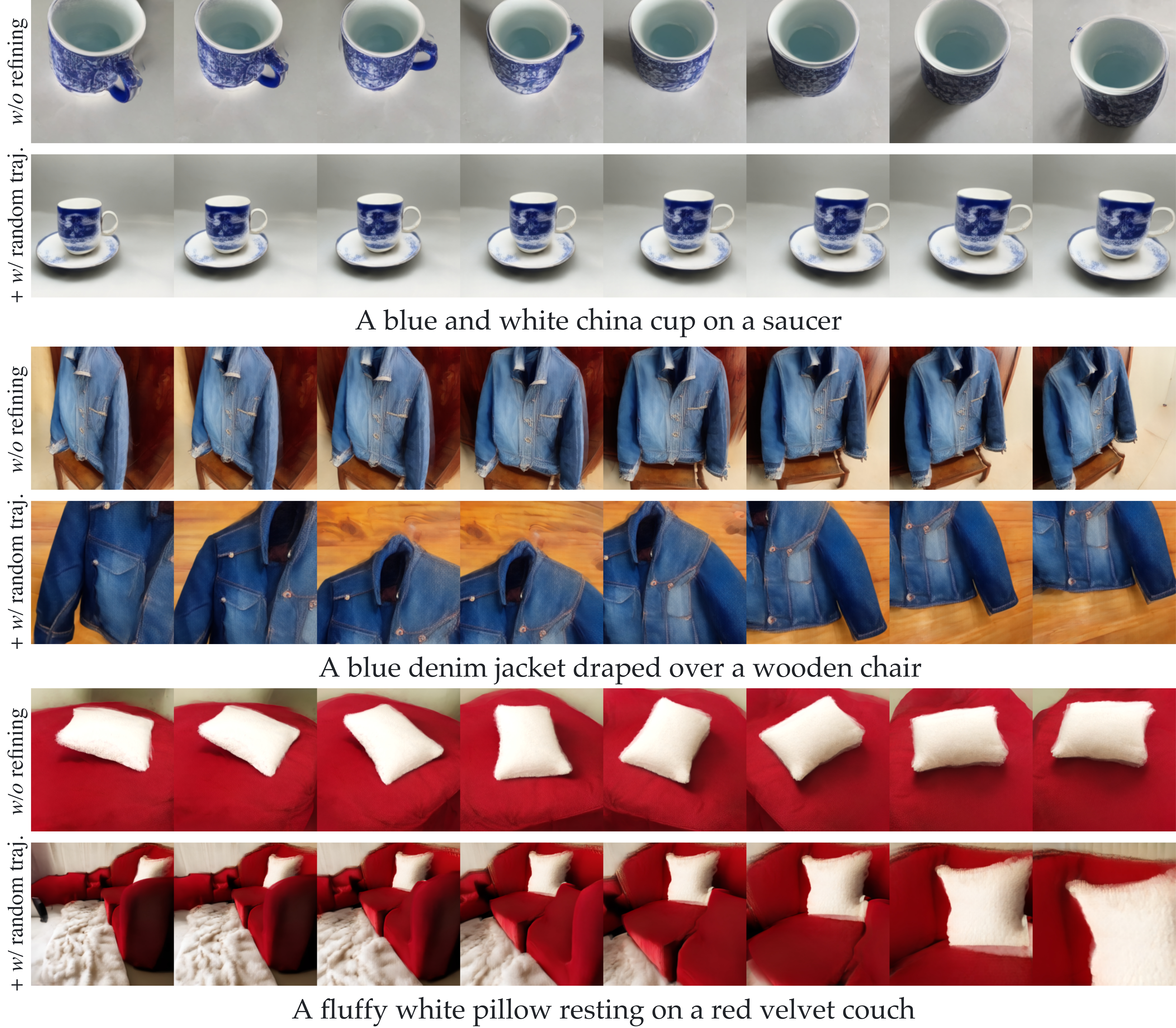}
  \caption{
  Generation results with randomly generated camera trajectories.
  }
  \label{fig.10}
\end{figure*}

In this experiment, we showcase the generation results with randomly generated camera trajectories for some object-level prompts in Fig~\ref{fig.10}. 
The results are generated by the GM-LDM without refinement.
It can be observed that without scene-specific camera trajectories, the generated image sequences are unsatisfying, with weird perspectives or limited camera range.
This ablation study further demonstrates the importance of scene-specific camera trajectories for our \modelname. 

\clearpage
\newpage

\end{document}